\documentclass[journal]{IEEEtran}
\usepackage{graphicx} 
\usepackage{german} 

\usepackage[backend=bibtex,style=ieee-alphabetic,natbib=true]{biblatex} 
\begin{document}
\title{Analyse der Entwicklungstreiber militärischer Schwarmdrohnen durch Natural Language Processing\\}
\author{Manuel~Mundt}
\maketitle

\begin{abstract}
Militärische Drohnen nehmen eine zunehmend stärkere Rolle in bewaffneten Konflikten ein, wobei der Einsatz mehrerer Drohnen im Schwarm sinnvoll sein kann. Wer die Treiber der Forschung sind und welche Subdomänen existieren wird in dieser Untersuchung durch NLP-Techniken auf der Basis von 946 Studien analysiert und visuell aufbereitet. Die meiste Forschung wird in der westlichen Welt durchgeführt, angeführt von den USA, dem Vereinigten Königreich und Deutschland. Durch Tf-idf-Scoring wird gezeigt, dass die Länder signifikante Unterschiede bei den untersuchten Subdomänen aufweisen. Insgesamt wurden in den Jahren 2019 und 2020 die meisten Werke veröffentlicht, wobei bereits 2008 ein signifikantes Interesse an militärischen Schwarmdrohnen bestand. Die vorliegende Studie gewährt einen ersten Einblick in die Forschung auf diesem Gebiet und gibt Anlass zu weiteren Untersuchungen.\\
\par Military drones are taking an increasingly prominent role in armed conflict, and the use of multiple drones in a swarm can be useful. Who the drivers of the research are and what sub-domains exist is analyzed and visually presented in this research using NLP techniques based on 946 studies. Most research is conducted in the Western world, led by the United States, the United Kingdom, and Germany. Through Tf-idf scoring, it is shown that countries have significant differences in the subdomains studied. Overall, 2019 and 2020 saw the most works published, with significant interest in military swarm drones as early as 2008. This study provides a first glimpse into research in this area and prompts further investigation.

\end{abstract}

\hfill 15. November 2022
\IEEEpeerreviewmaketitle

\section{Einleitung}
Bei Drohnen, oder auch UAV's handelt es sich um unbemannte Luftfahrzeuge, die automatisch per GPS an ihr Ziel navigieren oder alternativ ferngesteuert werden können und häufig mit Kameras ausgestattet sind. Dabei existieren Varianten mit starren Flügeln, wie bei Flugzeugen, oder drehenden Flügeln, wie bei Helikoptern. Genutzt werden sie im kommerziellen Bereich unter anderem zur Aufdeckung von Funklöchern, der geologischen Erschließung, der Fotografie, der Landwirtschaft oder zur Lieferung von Waren. \cite{ahirwar2019application} \\
\\
Neben der zivilen Nutzung können Drohnen allerdings auch (para-)militärisch eingesetzt werden \cite{ahirwar2019application, taha2019machine}. Anwendungsfelder sind dabei die Überwachung \cite{ahirwar2019application}, die Störung von Flugverkehr \cite{taha2019machine} oder der direkte Angriff durch abgefeuerte Flugkörper \cite{shaw2016predator}. Eine weitere Möglichkeit ist die Lenkung von, mit Sprengstoff beladenen, Drohnen in Ziele, bei der die Drohne selber geopfert wird \cite{gettinger2017loitering}.\\
\\
In den jüngsten militärischen Konflikten in Afghanistan, der Ukraine und zwischen Armenien und Aserbaidschan nehmen unbemannte Luftfahrzeuge eine zunehmend größere Rolle ein \cite{welt2021azerbaijan, shaw2016predator,antezza2022ukraine}. Unter anderem wurde dabei im Oktober 2022 erstmals eine Konfrontation zweier militärisch eingesetzter Drohnen dokumentiert, bei der eine ukrainische Drohne ein russisches Gerät zum Absturz brachte \cite{dronevsdrone2022}. Eine Neuheit ist die Nutzung von Drohnen, die im Schwarm eingesetzt werden. Mehrere, häufig vernetzte Drohnen, agieren dabei gemeinsam bei der jeweiligen Aufgabenstellung, etwa bei Suchaufträgen, der Aufspannung von Kommunikationsnetzwerken oder Paketlieferungen. \cite{alkouz2021provider} Auch auf dem militärischen Gebiet wird nach einer Suchanfrage auf \textit{Lens.org}\footnote{= ( drone OR UAV ) AND ( ( military OR ( armed OR combat ) ) AND swarm )} bereits seit 1986 vereinzelt geforscht. Vermehrt geforscht wird dabei ab 2003 mit einem starken Anstieg von 2012 bis 2019. Seit 2021 nimmt die Anzahl der veröffentlichten Arbeit wieder ab.\\
\\
Diese Studie soll die Forschung auf dem Gebiet der militärisch eingesetzten Schwarmdrohnen im zeitlichen Verlauf untersuchen. Dabei sollen unter anderem die relevanten Subthemen und die Treiber der Forschung identifiziert werden. Nach der Analyse werden die Ergebnisse (grafisch) ausgewertet und diskutiert.

\section{Untersuchungsgrundlage}
Grundlage für diese Studie sind wissenschaftliche Arbeiten, die den Begriff \textit{swarm} in Kombination mit \textit{combat}, \textit{armed} oder \textit{military} sowie mit \textit{uav} oder \textit{drone} in Titel oder Abstract haben. Dadurch werden Arbeiten erfasst, die sich mit Drohnen, bzw. UAVs beschäftigen, die in Schwärmen agieren und militärisch eingesetzt werden. \\
Als Quelle wurde \textit{Lens.org} gewählt, da hier Metadaten und Abstracts von passenden Arbeiten zu diesen Kriterien gefunden und heruntergeladen werden können. Um die aktuelleren Entwicklungen zu erfassen, wurde der Zeitraum seit 2003 für die Untersuchung gewählt. \\
\\
Heruntergeladen wurden 946 Datensätze, von denen jeder eine wissenschaftliche Arbeit repräsentiert. Um Duplikate in den Datensätzen zu entfernen, werden Titel und Abstract der Daten in eine Repräsentation aus Kleinbuchstaben überführt. Dopplungen in den ersten 40 Zeichen werden jeweils als Duplikate entfernt. Weiter werden Einträge herausgefiltert, die keinen Abstract haben, da dieser für die anschließende Analyse benötigt wird. Nach dem Prozess reduziert sich die Datenbasis auf 845 Datensätze.

\section{Datenanalyse}
Der erste Analyseschritt ist die Identifikation von Autorennetzwerken. Dazu werden die Autoren, die in den Rohdaten für jeden Datensatz per Komma getrennt vorliegen, normalisiert, wodurch eine Tabelle aus Autoren und Werken entsteht. Haben zwei Autoren an einem Werk gearbeitet, dann wird diese Zusammenarbeit gespeichert, was für alle Einträge durchgeführt wird. Es werden lediglich Beziehungen beibehalten, bei denen die Autoren mindestens 4 mal gemeinsam gearbeitet haben, da die Visualisierung ansonsten sehr unübersichtlich würde. Für jedes entstehende Autorennetzwerk wird jeweils bestimmt, in welchen Ländern die Arbeiten der Netzwerke veröffentlicht wurden. Diese Information wird für die Darstellung als Legende verwendet. Die Position der Knoten wird für die anschließende Darstellung durch den \textit{Fruchterman-Reingold}-Algorithmus\footnote{https://networkx.org/documentation/stable/reference/generated/ networkx.drawing.layout.spring \_layout.html} bestimmt, wobei die idealen Parameter empirisch ermittelt werden. Das visualisierte Netzwerk ist in Abbildung \ref{fig:aut_net} dargestellt.\\
\\
Um zu analysieren, welche Länder im zeitlichen Verlauf die meisten Veröffentlichungen aufweisen, werden zunächst die 5\% der Länder mit den wenigsten Publikationen aussortiert. Zudem werden alle Datensätze ohne Land entfernt, womit 342 Datensätze verbleiben. Als Visualisierung wird eine Heatmap erstellt, die in Abbildung \ref{fig:heat_land_jahr} dargestellt ist. \\
\\
Im nächsten Schritt soll für die bereits gezeigten Länder untersucht werden, woran geforscht wird. Dazu soll analysiert werden, welche Begriffe jeweils am wichtigsten sind.  Hierfür werden die Abstracts herangezogen und lemmatisiert, wobei nur Nomen, Verben und Adjektive berücksichtigt werden. Es wurde geprüft, ob hintereinander gereihte Nomen als zusammengesetzer Compound betrachtet werden sollten. Am Beispiel \textit{swarmsimulation} wird allerdings klar, dass die Einzelworte eine größere Bedeutung haben, da die Kombination nur selten auftritt.\\
Für die Lemmata werden anschließend die Tf-idf-Scores berechnet und in einer Term-Dokumenten-Matrix dargestellt. Durch dieses Verfahren wird für jedes Wort die \textit{Interessantheit} im Abstract kalkuliert \cite{sklearn_api}. Die entstehende Matrix wird je Land gruppiert, wobei die Aggregation der Scores durch die Bildung des Mittelwerts gewählt wird. Für die grafische Darstellung werden die 25 wichtigsten Begriffe genutzt und ihr Tf-idf-Score für jedes Land in einer Heatmap dargestellt. Einige dieser Begriffe sind dabei domänenunabhängig wie \textit{paper} oder \textit{solution} und wurden entfernt, da sie keine fachliche Bedeutung tragen. Zudem wurden die Synonyme \textit{Uav} und \textit{Drone} entfernt und fortan nur \textit{Drone} verwendet. Die finale Grafik ist in Abbildung \ref{fig:heat_begriff_land} dargestellt.\\
\\
Um Themenkomplexe der Domäne identifizieren zu können, sollen zunächst thematische Cluster gebildet werden, zu denen anschließend die wichtigsten Begriffe bestimmt werden sollen. Im ersten Schritt wird jeder Abstract in eine Vektor-Repräsentation überführt. In einem einfachen Ansatz können dazu alle Wörter in einem Abstract vektorisiert werden, wobei die Python-Bibliothek \textit{spaCy}\footnote{siehe \cite{spacy}} verwendet werden kann. Um dann einen Vektor pro Abstract zu erhalten, können alle Wort-Vektoren gemittelt werden. Ein weiterer Ansatz ist die Vektorisierung gesamter Abstracts durch die Nutzung von \textit{Transformer}-Modellen. Genutzt wird in dieser Studie die Python-Bibliothek \textit{Huggingface}\footnote{siehe \cite{huggingface}}, mit der das Modell \textit{all-MiniLM-L6-v2}\footnote{https://huggingface.co/sentence-transformers/all-MiniLM-L6-v2} eingesetzt wird, um Embeddings, also Vektoren, zu erhalten.\\
Um Cluster zu bilden, können verschiedene Algorithmen eingesetzt werden. Nahe liegt ein dichtebasierter Ansatz, der Cluster identifizieren kann, wenn diese eindeutig von einander unterscheidbar sind. \cite{sklearn_api} Um den optimalen Maximalabstand zweier Datensätze zu finden, bei denen sie noch dem selben Cluster zugewiesen werden, kann die \textit{Nächste-Nachbarn}-Methode verwendet werden \cite{dbscan}. Nach Finden und Anwenden des optimalen Parameters wird deutlich, dass lediglich ein einziger Cluster durch den dichtebasierten Algorithmus \textit{DBSCAN} identifiziert wird. Daher wird zurückgefallen auf den allgemein einsetzbaren Algorithmus \textit{KMeans}. \cite{sklearn_api} Durch die \textit{Elbow}-Methode\cite{syakur2018integration} wird \textit{3} als optimale Clusteranzahl identifiziert und ein etsprechendes Clustering durchgeführt. Für eine zweidimensionale Darstellung der Werke wird anschließend eine Dimensionsreduktion durchgeführt. Hierzu werden die Algorithmen \textit{t-sne} und \textit{pca} getestet \cite{anowar2021conceptual}. Nach einer Visualisierung wird die Reduktion per pca für die finale Darstellung gewählt, bei dem die Cluster gruppiert und in den meisten Punkten von einander getrennt vorhanden sind. \\
Analog zur letzten Darstellung werden hier für die drei Cluster die gemittelten Tf-idf-Scores für alle lemmatisierten und bereinigten Begriffe berechnet. Die fünf Begriffe, die für jeden Cluster am bedeutsamsten sind, werden als Legendenbeschriftung genutzt. Die dimensionsreduzierten Abstracts werden gefärbt nach ihrem Cluster als Streudiagramm in Abbildung \ref{fig:cluster_scatter} dargestellt.

\section{Ergebnisauswertung}

\begin{figure}
    \centering
\includegraphics[width=0.5\textwidth]{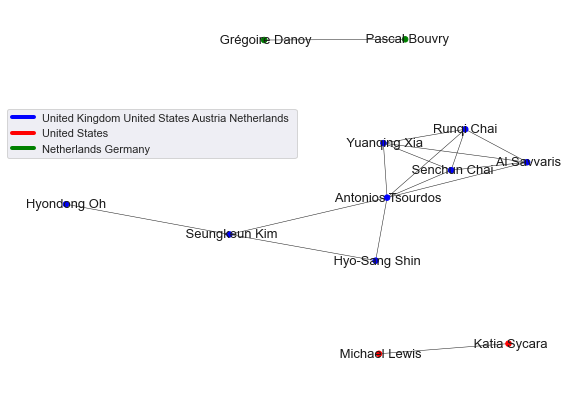}\\
    \caption{Autorennetzwerke mit mindestens 4 Zusammenarbeiten je Kante.}
    \label{fig:aut_net}
\end{figure}
In Abbildung \ref{fig:aut_net} sind Autorennetzwerke dargestellt, bei denen jede Kante mindestens 4 Zusammenarbeiten von zwei Autoren markiert. Neben zwei Autorenpaaren ist ein größeres Netzwerk aus insgesamt 8 Autoren sichtbar. Im Mittelpunkt steht dabei der Autor \textit{Antonios Tsourdos}, der mit jedem bis auf einen Autoren gearbeitet hat. Arbeiten dieses Netzwerkes wurden in den USA, im Vereinigten Königreich, Österreich und den Niederlanden veröffentlicht. Das erste Paar hat seine Werke lediglich in den USA publiziert und das letzte Paar veröffentlichte die Arbeiten in Deutschland und den Niederlanden. \\
Die Autorennetzwerke, die hier dargestellt sind, sind die bedeutendsten, die in den Daten zu finden sind. Dabei wurden alle Arbeiten dieser Autoren in der westlichen Welt veröffentlicht, was darauf schließen lässt, dass die Zusammenarbeit auf diesem Gebiet zwischen diesen Ländern am stärksten ist. Allerdings ist zu beachten, dass hier nur 12 der insgesamt 2498 Autoren in den Daten abgebildet sind. Es kann also lediglich ein Rückschluss auf die Zusammenarbeit, nicht aber über das tatsächliche Volumen der wissenschaftlichen Werke der Länder gezogen werden.\\
\\
\begin{figure}
    \centering
    \includegraphics[width=0.5\textwidth]{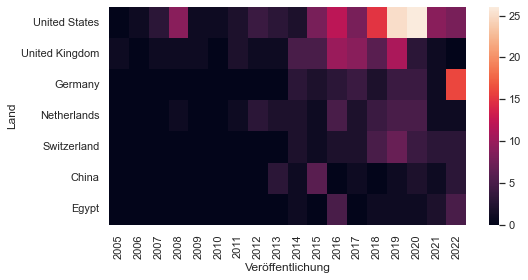}\\
    \caption{Heatmap der Publikationen je Land und Jahr.}
    \label{fig:heat_land_jahr}
\end{figure}
Um dies darzustellen, wurde die Heatmap in Abbildung \ref{fig:heat_land_jahr} erstellt. Zu erkennen sind hier die Anzahl der Publikationen für die Länder, die 95\% des Gesamtvolumens ausmachen. Je heller das Feld, desto mehr Veröffentlichungen gibt es. \\
Am deutlichsten sind die Jahre \textit{2019} und \textit{2020} in den \textbf{USA}, die jeweils über 20 Werke zählen. Insgesamt weisen die USA in vielen Jahren die meisten Arbeiten auf und sind damit der wichtigste Entwicklungstreiber. Bereits 2018 wurden hier mehr als 15 Werke über militärische Schwarmdrohnen herausgebracht. Direkt hinter den USA befindet sich das \textbf{Vereinigte Königreich}, wo vor allem von 2014 bis 2019 vermehrt geforscht wurde. Ebenfalls ein auffälliger Datenpunkt ist \textbf{Deutschland} im Jahr \textit{2022} mit etwa 20 Arbeiten, womit es in diesem Jahr die aktivste Nation ist. Es folgen die \textbf{Niederlande}, in denen von 2012 bis 2020 am meisten geforscht wurde. Die meisten Forschungen wurden in der \textbf{Schweiz} in 2018 und 2019 getätigt, wobei auch in den Jahren davor und danach Arbeiten veröffentlicht wurden. Als aktivstes Land aus der nicht-westlichen Welt befindet sich \textbf{China} auf dem sechsten Rang. Die meisten Publikationen wurden 2013, 2015 und 2022 getätigt. Auf Rang 7 befindet sich schließlich \textbf{Ägypten}, das 2016 und 2022 die meisten Veröffentlichungen aufweist.\\
Auch in dieser Grafik zeigt sich die Dominanz der westlichen Welt in der Erforschung dieser Domäne. Limitierend wirkt jedoch, dass lediglich 360, der in den Rohdaten vorhandenen, Datensätze über ein Land verfügen, wodurch das Bild verfälscht werden könnte. Etwa wenn die Verteilung der Länder in den unvollständigen Datensätzen abweicht.\\
\\
\begin{figure}
    \centering
    \includegraphics[width=0.5\textwidth]{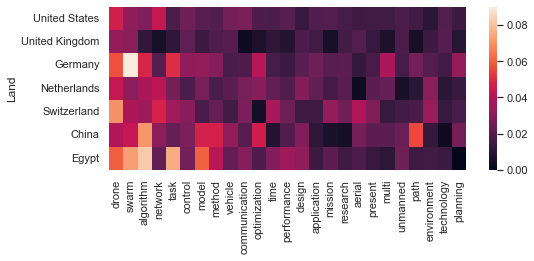}\\
        \caption{Heatmap der wichtigsten Begriffe je Land.}
    \label{fig:heat_begriff_land}
\end{figure}
Ein weiteres Ziel dieser Arbeit ist die Identifikation von Themen innerhalb der Domäne. In Abbildung \ref{fig:heat_begriff_land} sind dazu die 25 wichtigsten Begriffe der Abstracts dargestellt und nach ihrer gesamten Wichtigkeit sortiert. Auf der X-Achse befinden sich wieder die Länder, die kumuliert 95\% der Arbeiten ausmachen. Je heller das Feld hier dargestellt ist, desto wichtiger ist der Begriff für dieses Land. Für die oberen Länder sind mehr Abstracts und somit mehr Begriffe vorhanden, da hier die meisten Arbeiten veröffentlicht wurden. Begriffe, die in einem Abstract wichtig sind, sind in anderen möglicherweise weniger interessant. Dadurch, dass die Tf-idf-Scores gemittelt werden, sind die Ausprägungen daher weniger prägnant, was die gedämpfteren Farben in den USA und dem Vereinigten Königreich erklärt.\\
\\
Für die \textbf{USA}, die am meisten Publikationen aufweisen, sind die vier wichtigsten Begriffe auch die insgesamt wichtigsten. \textit{Drone} und \textit{Network} haben hier die höchsten Scores, gefolgt von \textit{Swarm} und \textit{Algorithm}. Die Forschung scheint hier einen Schwerpunkt auf Netzwerke und die benötigten Algorithmen zu legen. Ein weiterer heller Punkt ist \textit{Communication}, was dafür sprechen könnte, dass an der Kommunikation der Drohnen untereinander gearbeitet wird. Weniger stark ausgeprägt sind die Wichtigkeiten der Begriffe im \textbf{Vereinigten Königreich}. Neben \textit{Drone} und \textit{Swarm} sind \textit{Control} und \textit{Vehicle} hier am interessantesten. Es könnte ein Indiz dafür sein, dass hier vermehrt an der Steuerung der Luftfahrzeuge gearbeitet wird. In \textbf{Deutschland} liegt der größte Fokus auf dem Begriff \textit{Swarm}, doch auch \textit{Drone}, \textit{Algorithm}, \textit{Task}, \textit{Optimization}, \textit{Multi} und \textit{Planning} erscheinen hell in der Grafik. Hier wird scheinbar an einer großen Bandbreite an Themen gearbeitet. In den \textbf{Niederlanden} gibt es weniger starke Ausprägungen als in Deutschland, wobei sich die 4 wichtigsten Begriffe mit den insgesamt wichtigsten decken. Weiter sind hier die Worte \textit{Optimization}, \textit{Design} und \textit{Environment} im Fokus. Ob die Begriffe im Zusammenhang stehen, oder einzelne Subthemen darstellen, kann aufgrund dieser Auswertung jedoch nicht bestimmt werden. Die \textbf{Schweiz} legt ihren Fokus laut der Grafik ähnlich wie die USA, sodass \textit{Drone} und \textit{Network} die wichtigsten Begriffe sind. Ob es sich dabei um Netzwerke von Drohnen oder Kommunikationsnetze handelt, kann so allerdings nicht bestimmt werden. Begriffe, in denen sich die Schweiz am meisten zu den restlichen Ländern differenziert, sind außerdem \textit{Mission}, \textit{Research}, \textit{Aerial} und \textit{Present}. \\
Als erstes Land außerhalb der westlichen Welt haben in \textbf{China} die Begriffe \textit{Algorithm} und \textit{Path} die höchste Bedeutung. Geforscht wird somit vermutlich nach Algorithmen für die Wegfindung der Drohnen. \textit{Model} und \textit{Method} sind zudem Wörter, welchen in chinesischen Publikationen eine hohe Bedeutung zukommt. Ein ebenfalls hoher Fokus auf Algorithmen, liegt in \textbf{Ägypten}, wo außerdem die Begriffe \textit{Task} und \textit{Swarm} ausgeprägt sind.\\
Diese Analyse kann einen ersten Einblick darin geben, welche Forschungszweige in der Forschung der betrachteten Länder von besonderem Interesse sind. Es kann jedoch nicht gesagt werden, ob diese Begriffe stets in Zusammenhang zueinander stehen. Dazu könnten weiterführende Analysen genutzt werden.\\
\\
\begin{figure}
    \centering
\includegraphics[width=0.5\textwidth]{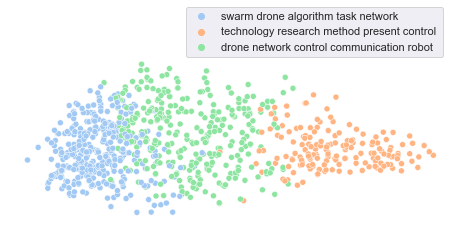}\\
    \caption{Streudiagramm der geclusterten Abstracts.}
    \label{fig:cluster_scatter}
\end{figure}
Um voneinander abgrenzbare Themenbereiche zu identifizieren, wurde eine Clusteranalyse der Abstracts durchgeführt und diese im Streudiagramm Abbildung \ref{fig:cluster_scatter} visualisiert. Hier zeigt sich, dass die gewählte Methode keine eindeutig separierbaren Cluster hervorbringt, auch wenn im blauen und im orangenen Cluster auffällige Häufungen vorhanden sind. Den drei Clustern wurden jeweils die 5 wichtigsten Begriffe zugeordnet. Der blaue Cluster hat neben den Begriffen \textit{Swarm} und \textit{Drone} als wichtigste Worte \textit{Algorithm}, \textit{Task} und \textit{Network}, was die 5 insgesamt wichtigsten Begriffe der vorherigen Analyse sind. Die Arbeiten im grünen Cluster setzen sich ebenfalls mit Netzwerken auseinander und beschäftigen sich den Begriffen nach mit Robotik, Kommunikation und der Kontrolle der Fahrzeuge. Eher methodenorientiert sind scheinbar die Werke des orangenen Clusters, da neben \textit{Control} keine Begriffe auf spezifische Themen hinweisen. Vielmehr sind die wichtigsten Worte dieser Abstracts \textit{Technology}, \textit{Research}, \textit{Method} und \textit{Present}.

\section{Fazit \& Ausblick}
In der Einleitung wurde gezeigt, dass militärische Drohnen eine immer stärkere Rolle in Konflikten spielen, wobei es verschiedene Einsatzzwecke gibt, bei denen auch der Einsatz mehrerer vernetzter Fahrzeuge im Schwarm sinnvoll sein kann.\\
Die Relevanz der Thematik wird durch die Anzahl der aktuellen Publikationen untermauert. Welche Subdomänen hier existieren und wer die Treiber der Forschung sind, war Untersuchungsgegenstand dieser Studie.\\
\\
Um diese Themen beantworten zu können, wurden 946 Datensätze über wissenschaftliche Veröffentlichungen seit 2003 heruntergeladen und analysiert. Dabei wurden Netzwerke von häufig zusammenarbeitenden Autoren identifiziert und die Länder, in denen sie publizieren, erfasst. Die größten Netzwerke sind stets innerhalb der westlichen Welt angesiedelt, was als Indiz dafür gilt, dass die Zusammenarbeit in der Domäne hier am stärksten ist.\\
Weiter wurden die wichtigsten Länder mit den meisten Publikationen ermittelt, wobei die Länder der ersten 5 Ränge in der westlichen Welt liegen, was dafür spricht, dass hier am meisten an militärischen Schwarmdrohnen geforscht wird. Zu erkennen ist ebenfalls, dass die Forschung 2019 und 2020 ihren Höhepunkt fand und seitdem rückläufig ist.\\
Für die wichtigsten Länder wurde anschließend per Tf-idf Verfahren berechnet, welche Begriffe innerhalb der Abstracts am wichtigsten sind, um zu erfassen, welche Themenbereiche jeweils am bedeutendsten sind. Hier zeigt sich, dass es signifikante Unterschiede zwischen den Nationen gibt.\\
In der Clusteranalyse der Abstracts wurde herausgefunden, dass die Arbeiten nicht eindeutig voneinander separierbar sind, wenngleich es thematische Häufungen gibt. Hierbei werden zwischen den Clustern wieder signifikante thematische Unterschiede deutlich.\\
\\
Die vorliegende Studie bietet einen ersten Überblick über die Entwicklung und die Themen innerhalb der Domäne militärischer Schwarmdrohnen. Damit wird ein Einstieg für tiefergehende Analysen geboten. Limitierend wirkt für diese Arbeit in erster Linie die Datengrundlage. \textit{Lens.org} bietet zwar eine komfortable, aber vermutlich unvollständige Basis, die mit weiteren Quellen angereichert werden sollte, was größeren Aufwand im Preprocessing verursachen würde. Zudem sind die heruntergeladenen Daten teilweise unvollständig gefüllt. Nur 360 Datensätze haben so ein gefülltes Land. Ebenfalls unvollständig sind die Abstracts. Durch die fehlenden Daten könnten die Darstellungen dieser Studie verfälscht oder wichtige Entwicklungen nicht erfasst worden sein. Kritisch betrachtet werden müssen die Resultate auch, da gerade auf dem Bereich der militärischen Forschung nicht alle Forschungen veröffentlicht werden.\\
Es muss ebenfalls hinterfragt werden, ob die gewählten Methoden zu den bestmöglichen Resultaten geführt haben. Denkbar wäre etwa der vermehrte Einsatz von Transformern zur Informationsextraktion. Für die Cluster des Streudiagramms wäre es beispielsweise möglich, Zusammenfassungen oder weiterführende Analysen der Zentroide zu erstellen, die stellvertretend für den jeweiligen Cluster stünden. Weiter wäre es möglich, verschiedene Techniken des Clusterings und der Dimensionsreduktion empirisch zu testen. \\ 
\\
Aufbauend auf dieser Arbeit könnten tiefergehende Analysen mit den eben vorgestellten Ansätzen durchgeführt werden. Ebenfalls könnten für die identifizierten Themenkomplexe Metastudien angefertigt werden, die einen detaillierteren Einblick gewähren können. Schließlich wäre es interessant zu untersuchen, welche Subdomäne das Thema \textit{Drohne} aktuell dominiert, da die Forschung über Schwarmdrohnen, wie gezeigt, seit 2021 rückläufig ist.

\printbibliography
\end{document}